\documentclass{article}

\usepackage{spconf}
\usepackage{amsmath}
\usepackage{graphicx}
\usepackage{bm}
\usepackage{multirow}
\usepackage{amsfonts}
\usepackage{xcolor}
\usepackage{booktabs}
\usepackage{xspace}
\usepackage{tabularx}
\usepackage{amssymb}
\usepackage{float}
\usepackage{verbatim}
\usepackage[
    colorlinks=true,
    citecolor={[rgb]{0.98,0,0}},
    linkcolor={[rgb]{0.98,0,0}},
]{hyperref}
\usepackage{arydshln}
\usepackage{color, colortbl}
\usepackage{tikz}
\usepackage{pifont}
\usepackage[symbol]{footmisc}
\usepackage{tablefootnote}
\usepackage{hyperref}
\usepackage{adjustbox}

\def\x{{\mathbf x}}
\def\z{{\mathbf z}}
\def\p{{\mathbf p}}
\def\y{{\mathbf y}}
\def\M{{\mathbf M}}

\definecolor{LightCyan}{rgb}{0.88,1,1}

\newcommand{\cmark}{\ding{51}}
\newcommand{\xmark}{\ding{55}}

\title{TAMFormer: Multi-Modal Transformer with Learned Attention Mask for Early Intent Prediction}

\name{Nada Osman, Guglielmo Camporese, Lamberto Ballan}
\address{Department of Mathematics “Tullio Levi-Civita”, University of Padova, Italy \\
\tt\footnotesize{\{nadasalahmahmoud.osman, guglielmo.camporese\}@phd.unipd.it}\\
\tt\footnotesize{lamberto.ballan@unipd.it}}

\begin{document}
%\ninept
\maketitle

%%% abstract
\begin{abstract}
Human intention prediction is a growing area of research where an activity in a video has to be anticipated by a vision-based system. To this end, the model creates a representation of the past, and subsequently, it produces future hypotheses about upcoming scenarios.
In this work, we focus on pedestrians' early intention prediction in which, from a current observation of an urban scene, the model predicts the future activity of pedestrians that approach the street. Our method is based on a multi-modal transformer that encodes past observations and produces multiple predictions at different anticipation times. Moreover, we propose to learn the attention masks of our transformer-based model (\textbf{T}emporal \textbf{A}daptive \textbf{M}ask Trans\textbf{former}) in order to weigh differently present and past temporal dependencies. We investigate our method on several public benchmarks for early intention prediction, improving the prediction performances at different anticipation times compared to the previous works.
\end{abstract}


%%% introduction
\vspace{-2mm}
\section{Introduction}
\label{sec:intro}
In the last years, the development of computer vision algorithms has seen a massive improvement thanks to the advent of deep learning enabling new applications in the context of autonomous driving, video surveillance, and virtual reality.
The visual understanding capabilities of deep learning models have been adopted in various domains, from smart cameras used in video surveillance to cognitive systems in robotics and multi-modal sensors for autonomous driving. Moreover, a recent interesting direction involves predicting future activities that can be anticipated from a visual content~\cite{Furnari2019WhatWY, Becattini2021, Girdhar2021AnticipativeVT}. Some applications enabled by the models designed for action anticipation are pedestrian intention prediction from a smart camera and ego-centric action anticipation from a robotic agent. In this work, we investigate the early intention prediction of pedestrians in an urban environment. In particular, \textbf{\textit{i)}} we propose a new model for early intent prediction based on a multi-modal transformer; \textbf{\textit{ii)}} we propose a new mechanism for learning the attention masks inside the transformer that leads to better performances and more efficient computation; and \textbf{\textit{iii)}} we conduct several experiments and model ablations on different datasets obtaining state-of-the-art results on the early intent prediction task.

%%% related works
\vspace{-2mm}
\section{Related Works}
\label{sec:related_works}
\vspace{-1mm}

\noindent 
\textbf{Action Recognition.} Video action recognition is a well investigated problem that, in recent years, has experienced massive improvements thanks to the recent progress of deep learning. Specifically, traditional hand-crafted video approaches~\cite{Efros2003RecognizingAA, Klser2008ASD, Laptev2008LearningRH, Ballan2012, Peng2014ActionRW} have been replaced by models based on recurrent neural networks~\cite{Donahue2015LongtermRC, Jiang2019STMSA, Li2018RecurrentTP, Li2018VideoLSTMCA, Ng2015BeyondSS}, 2D CNN~\cite{Wang2015ActionRW, Wang2016TemporalSN, Wu2018CompressedVA}, and 3D CNN~\cite{Carreira2018ASN, Feichtenhofer2020X3DEA, Feichtenhofer2019SlowFastNF, Girdhar2019VideoAT, Li2018VideoLSTMCA, Qiu2017LearningSR}. Transformers~\cite{Vaswani2017AttentionIA} have been also investigated for spatio-temporal modeling~\cite{Fan2021MultiscaleVT, Arnab2021ViViTAV, Girdhar2019VideoAT} improving the state-of-the-art performances on video related problems including video action recognition.

\smallskip
\noindent 
\textbf{Action Anticipation and Intent Prediction.} Recently, anticipating actions on videos gained attention given the development of new methods~\cite{Furnari2019WhatWY, Sener2020TemporalAR, Girdhar2021AnticipativeVT, Wu2022MeMViTMM, Camporese2021KnowledgeDF}, datasets~\cite{Damen2020RescalingEV, Rasouli2019PIEAL, Rasouli2017AreTG}, and applications such as autonomous driving, human-robot interaction and virtual reality. In particular, in urban environments, the pedestrian intent prediction from third-view cameras is a growing area~\cite{Rasouli2019PIEAL, Rasouli2017AreTG} in which models are designed to predict the future activity of pedestrians. 

\smallskip
\noindent 
\textbf{Temporal Modeling on Vision Problems.}
Video-based models need to process spatial and temporal information. Usually, the temporal axis is considered an independent component of the video, and in the model design, the spatial information is processed differently from the temporal one. Recent works proposed to model the temporal at different frame rates~\cite{Feichtenhofer2019SlowFastNF}, with multiple time-scales~\cite{Sener2020TemporalAR, Osman2021SlowFastRL}, and with an adaptive frame-rate~\cite{Wu2019AdaFrameAF}. However, in such works, the temporal sampling strategy of the frames is fixed and treated as a hyper-parameter of the model. For this reason, in this work, we explore and propose an adaptive mechanism for weighting the importance of the current and past frames by learning the attention mask inside our transformer model.

%%% methodology
%\vspace{-2mm}
\begin{figure*}[!ht]
    \centering
    \includegraphics[width=.95\linewidth]{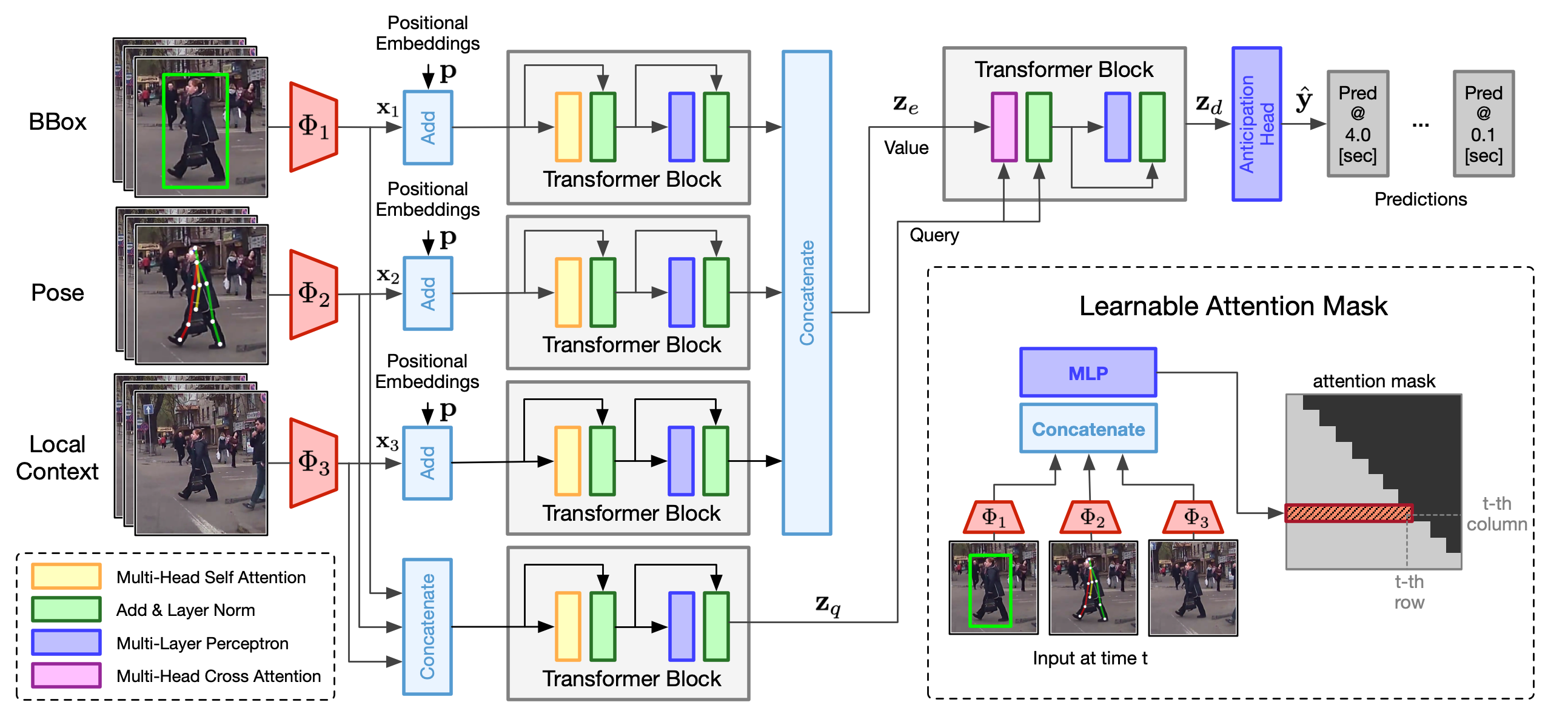}
    \caption{Our TAMformer model architecture.}
    \label{fig:model}
\end{figure*}

\section{Our Method}
\label{sec:method}

Our proposed TAMformer model, depicted in Fig.~\ref{fig:model}, has three major components: the \textit{Encoding} in which the multi-modal input is encoded, the \textit{Query} where the future query is built, and the \textit{Decoding} where the future prediction is computed.

\smallskip
\noindent
\textbf{Encoding.} 
%A transformer block $TF^E_m$ is applied to the $m$-th modality temporal sequence extracted by $\Phi_m$, illustrated in Fig.~\ref{eq:encoding}, where $LN$ is the Layer Norm. 
%Each input modality $\x_m \in \mathbb{R}^{T \times D_m}$, where $m \in \{1, \dots, M\}$, is linearly projected with $\Phi_m$ to a fixed size embedding and subsequently passed to a transformer block $TE_m$ that creates an encoded representation $\z_m \in \mathbb{R}^{T\times D_m}$:
Raw images are projected to different modalities with $\Phi_m$, where $m \in \{1, \dots, M\}$, to $\x_m \in \mathbb{R}^{T \times D_m}$ and subsequently passed to a transformer block $TE_m$ that creates an encoded representation $\z_m \in \mathbb{R}^{T\times D_m}$:

% \vspace{-0.2cm}
% \begin{equation}
%     \begin{array}{c}
%         A^E_m = LN(MHSA(\Phi_m(x_m))+\Phi_m(x_m))\\
%         TF^E_m = LN(MLP(A^E_m)+A^E_m)
%     \label{eq:encoding}
%     \end{array}
% \end{equation}
\vspace{-4mm}
\begin{equation*}
        \z_m = TE_m(\x_m + \p), \quad \z_e = Cat\Big[ \z_1, \dots, \z_M \Big].
    \label{eq:encoding}
\end{equation*}

\noindent
In order to preserve the order of the sequence, the positional encodings $\p$ are added to the input sequence after the linear projection.
Our model does not assume any particular input modalities, however in our work we used the RGB local context, bounding box coordinates, pose, and the vehicle speed.

\smallskip
\noindent
\textbf{Query.} Instead of applying a single late fusion of the encoded sequences, we allow the input features to interact in an early fusion step. A transformer block $TQ$ processes the concatenated features, creating a query at each time step, as follows: 

% \vspace{-4mm}
% \begin{equation}
%     \begin{array}{c}
%         A^Q = LN(Attention(\underset{m}{\oplus} \, \Phi_m)+\underset{m}{\oplus} \, \Phi_m)), \\
%         TF^Q = LN(MLP(A^Q)+A^Q)
%     \end{array}
%      \label{eq:query}
% \end{equation}
\vspace{-1mm}
\begin{equation*}
    \tilde{\z}_q = Cat\Big[ \x_1, \dots, \x_M \Big], \quad \z_q = TQ(\tilde{\z}_q)
    \label{eq:query}
\end{equation*}

%\smallskip
\vspace{-1mm}
\noindent 
\textbf{Decoding.} 
A transformer decoder block $TD$ processes the encoded representation $\z_e$ and for each query in $\z_q$ produces a decoded representation through the cross-attention mechanism that subsequently is projected to the final prediction as follows:
% A transformer decoder block $TD$ is used (\ref{eq:decoding}). The input values to decoding are the lately fused encodings and the queries extracted by (\ref{eq:query}). Finally, the anticipated intention at step $i$ is given by $P_i = Sigmoid(MLP(TF^D_i))$.

% \vspace{-0.5cm}
% \begin{equation}
%   \begin{array}{c}
%         A^D = LN(Attention(\underset{m}{\oplus} \, TF^E_m,TF^Q)+TF^Q)\\
%         TF^D = LN(MLP(A^D)+A^D)
%     \end{array}
%     \label{eq:decoding}
% \end{equation}

\vspace{-4mm}
\begin{equation*}
    \z_d = TD(\z_e, \z_q), \quad \hat{\y} = Sigmoid(MLP(\z_d))
    \label{eq:decoding}
\end{equation*}

%\smallskip
\noindent 
\textbf{Learning Attention Masks.} Usually, video frames are redundant when processed at a high frame rate, and, by contrast, at a low frame rate, the information can be lost as the sampling does not consider frame importance. For these reasons, we propose a method that allows the model to choose the frames that maximize the information and minimize redundancy. As depicted in Fig.~\ref{fig:model}, at the $t$-th step, the input features are concatenated and fed to a feed-forward network that outputs a learned mask $\M_t$. We decided to encode the representations at a full frame rate (30 FPS) and to make predictions at a sub-sampled frame rate (10 FPS) for more efficient computation. In our model, we have two types of masks $\M_e$, $\M_d$ related to the encoding and decoding transformer blocks and, in order to avoid future information conditions in the present prediction, the masks are causal, and their $t$-th rows are predicted as follows:
% Precisely, the model has $3$ types of transformer: $TF^E$ is allowed to observe the whole $4.5$s sequence ($136$ frames), while $TF^Q$ and $TF^D$ are adapted to the downsampled predictions at $10$ pred./s. Accordingly, we have two learnable masks: $M^E_{(L_E \times L_E)}$ and $M^D_{(L_D \times L_D)}$, where $L_E=136, L_D=46$. To ensure causality, the model only learns $M^{E,D}_i$ until $i$, while masking the future part after $i$ with zeros, as in (\ref{eq:mask}).

\vspace{-1mm}
\begin{equation*}
    \begin{array}{c}
    \tilde{\z} = Cat\big[ \x_1, \dots, \x_M \big], \\ 
    \M_{[:t]} = Sigmoid(MLP(\tilde{\z}_{[:t]})).
    \label{eq:mask}
    \end{array}
\end{equation*}

\begin{table*}[!ht]
\centering
  \begin{adjustbox}{width=\linewidth,center}
  \setlength{\tabcolsep}{4pt}
     \begin{tabular}{l c c c c c c c c c c c c c c c c c c}
        \toprule
        &  \multirow{2}{*}{\begin{tabular}{@{}c@{}} Visual \\ Backbone\end{tabular}} & 
        \multirow{2}{*}{Blocks} & 
        \multirow{2}{*}{Fusion} & 
        \multirow{2}{*}{$t_a$} & \multirow{2}{*}{$S_{temp}$} &
        &\multicolumn{3}{c}{\textbf{PIE}}
        && \multicolumn{3}{c}{\textbf{JAAD}$_{all}$}
        && \multicolumn{3}{c}{\textbf{JAAD}$_{beh}$}\\
          \cmidrule{8-10} \cmidrule{12-14} \cmidrule{16-18}
         & & & & & & & Acc & AUC & F1 & & 
         Acc & AUC & F1 & &
         Acc & AUC & F1\\
         \cmidrule{2-6} \cmidrule{8-10} \cmidrule{12-14} \cmidrule{16-18}
         %\gray{PCPA~\cite{Kotseruba2021BenchmarkFE}} & \multirow{2}{*}{C3D}&\multirow{2}{*}{GRU}&\multirow{2}{*}{L-ATT}& \multirow{2}{*}{$[2-1]$s} & \multirow{2}{*}{$30$FPS} && \gray{$0.87$} & \gray{$0.86$} & \gray{$0.77$} & & \gray{$0.85$} & \gray{\textbf{0.86}} & \gray{\textbf{0.68}} & & \gray{$0.58$} & \gray{$0.5$} & \gray{$0.71$} \\
         PCPA\tablefootnote{\label{foot:reproducibility}The reported results are our run of PCPA. As mentioned in the following \href{https://github.com/ykotseruba/PedestrianActionBenchmark/issues/15}{GitHub issue}, there are some issues on reproducing the results of the original paper from the code.}~\cite{Kotseruba2021BenchmarkFE} & C3D & GRU & L-ATT & $[2-1]$s & $30$ FPS & & $0.86$ & \textbf{0.86} & $0.77$ & & $0.8$ & $0.79$ & $0.57$ & & $0.56$ & $0.5$ & $0.67$\\
         \cdashline{1-18}\noalign{\vskip 0.5ex}
         R-LSTM~\cite{Furnari2019WhatWY} & \multirow{3}{*}{VGG16} & \multirow{3}{*}{LSTM}& \multirow{3}{*}{L-ATT}& \multirow{3}{*}{$[4-0.1]$s} & \multirow{3}{*}{$10$ FPS} && $0.76$ & $0.67$ & $0.52$ & & $0.86$ & $0.76$ & $0.6$ & & $0.65$ & $0.59$ & $0.74$\\
         RU-LSTM~\cite{Furnari2019WhatWY} &&&&&&&  $0.87$ & $0.84$ & $0.77$ & & $0.86$ & $0.78$ & $0.62$ & & $0.69$ & $0.62$ & $0.78$\\
         G-RULSTM\cite{Osman2022EarlyPI} &&&&&&& \textbf{-} & \textbf{-} & \textbf{-} & & $0.86$ & $0.8$ & $0.63$ & & $0.72$ & $0.65$ & $0.8$\\
         \cdashline{1-18}\noalign{\vskip 0.5ex}
         TAMformer (ours) & VGG16&TF&EC+LC & $[4-0.1]$s & Adaptive && \textbf{0.88} & \textbf{0.86} & \textbf{0.79} & & \textbf{0.88} & \textbf{0.83} & \textbf{0.68} & & \textbf{0.73} & \textbf{0.69} & \textbf{0.8}\\
         \bottomrule
    \end{tabular}
    \end{adjustbox}
    \caption{Comparing different SOTA models in the standard anticipation range $[2-1]$s}
    \label{tab:sota}
\end{table*}

% \smallskip
% \vspace{1mm}
\noindent  
\textbf{Auxiliary Loss.} Typically, anticipation models perform better as they get closer to the anticipated action. Thus, we propose an auxiliary regularization loss function:

\vspace{-1mm}
\begin{equation*}
    \mathcal{L}_r = \sum_t\| \z_d[t] - \z_d[T] \|^2
    \label{eq:lw}
\end{equation*}

\vspace{-1mm}
\noindent 
that minimizes the gap between the current decoder embedding $\z_d[t]$ and the final one $\z_d[T]$. We found beneficial to train the model in two stages: we first pre-train the system using only the cross-entropy loss $\mathcal{L}_{ce}$ for action anticipation, and subsequently we add the regularization term $\mathcal{L}_r$ to the total loss ($\mathcal{L}=\mathcal{L}_{ce}+\mathcal{L}_{r}$), encouraging the earlier anticipation predictions to benefit from the last decoder representation that can observe the whole sequence before the action starts.

% \vspace{-0.2cm}
% \begin{equation}
%   \mathcal{L}_w = \frac{\sum^{L_e}_{k=1} (TF^D_{i,k}-TF^D_{-1,k})^2}{L_e}
%     \label{eq:lw}
% \end{equation}

%\smallskip
\noindent 
\textbf{Data Augmentation.} In contrast to the standard protocol, \cite{Kotseruba2021BenchmarkFE}, we abandon overlapped samples and follow the proposed protocol in \cite{Osman2022EarlyPI}, treating each pedestrian as a single sample. Consequently, a hard reduction in the number of samples is present, compared to \cite{Kotseruba2021BenchmarkFE}. However, transformers require large training data for the best results. Accordingly, we propose a data augmentation procedure to increase the training data. As in \cite{Osman2022EarlyPI}, the observation length is $4.5$s, ignoring any earlier frames in the sample. We benefit from such frames to augment the samples, replacing the encoding window with earlier frames when they exist. Thus, more versions of the same sample with different encoding windows are available.

%%% experiments
%\vspace{-2mm}
\section{Experimental Results}
\label{sec:results}

\noindent 
\textbf{Datasets and Metrics.} We evaluate our method on JAAD~\cite{Rasouli2017AreTG} and PIE~\cite{Rasouli2019PIEAL} datasets. JAAD contains two subsets: JAAD$_{beh}$ with only behaviorally annotated subjects ($495$ crossing and $191$ not crossing), and JAAD$_{all}$ with an additional $2100$ not crossing samples. Conversely, PIE contains $1842$ behaviorally annotated pedestrians ($519$ crossing and $1323$ not crossing), in addition to more annotations of the ego-vehicle, i.e., speed. Following~\cite{Kotseruba2021BenchmarkFE}, we evaluate the models with the standard classification metrics: Accuray, AUC, and F1-Score.

\smallskip
\noindent 
\textbf{Implementation Details.} The training procedure includes two phases (500 epochs each): a pre-training phase on action antipation and a tuning phase with the regularizer $\mathcal{L}_r$. We used the SGD optimizer with learning rates $lr = \{10^{-5}, 10^{-2}, 10^{-3}\}$ for PIE, JAAD$_{all}$, and JAAD$_{beh}$ respectively. Each transformer block has $N_h=6$ heads, $ff_{dim}=1024$, and the $MLP$ producing the learned masks consists of $N_l=3$ layers with sizes $\{128, 64, 32\}$. %After training, we find the the best classification threshold, given the problematic imbalance in the datasets towards the negative class (not crossing). 

\smallskip
\noindent 
\textbf{Results.} We compare our model with PCPA~\cite{Kotseruba2021BenchmarkFE}, that represents the SOTA work in intent prediction and an adapted PCPA version that can produce earlier anticipations. Although we are not applying the overlapping protocol in \cite{Kotseruba2021BenchmarkFE}, we align with it on the used samples and anticipation range during evaluation to allow for a fair comparison. Additionally, we compare with a single LSTM (R-LSTM), RULSTM~\cite{Furnari2019WhatWY}, and G-RULSTM~\cite{Osman2022EarlyPI}. Following \cite{Kotseruba2021BenchmarkFE}, Table~\ref{tab:sota} reports the comparison in the anticipation range of $[2-1]$ s, and the main architecture differences. We observe a F1-score out-performance gap that reaches $+2\%$ on PIE and $+5\%$ on JAAD$_{all}$, comparing our TAMformer to the best model in the table. Moreover, we reported a comparison on different anticipation times from $4$s to $1$s in Table~\ref{tab:overtime} and, depending on the dataset, we notice two trends: for PIE, TAMformer outperforms by almost $+2\%$ on F1-score in all anticipation times. Nevertheless, on JAAD, our model suffers a degraded performance at early anticipation ($[4 - 3]$s) while maintaining the improvements on JAAD$_{all}$ (maximum $+9\%$)  and on JAAD$_{beh}$ (maximum $+2\%$). The reduction in training samples in early anticipation ($>50\%$ on JAAD) could explain this degradation as transformers need lots of training samples.

\begin{figure}[!ht]
    \centering
    \includegraphics[width=\linewidth]{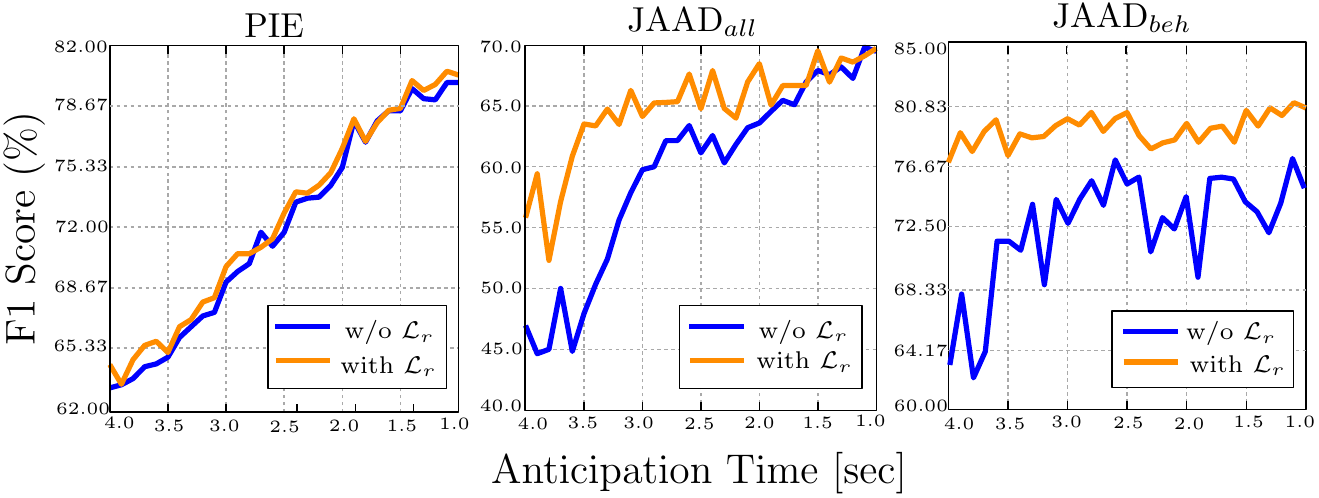}
    \caption{Effect of using the $\mathcal{L}_r$ loss}
    \label{fig:lw_evaluation}
\end{figure}

% \begin{figure}[!ht]
%     \centering
%     \begin{subfigure}[b]{\linewidth}
%         \centering
%         \caption{PIE}
%         \includegraphics[width=0.95\linewidth]{images/pie}
%         \label{fig:lw_evaluation_pie}
%     \end{subfigure}
%     %\vspace{4mm}
%     \begin{subfigure}[b]{\linewidth}
%         \centering
%         \caption{JAAD\textsubscript{all}}
%         \includegraphics[width=0.95\linewidth]{images/jaad_all}
%         \label{fig:lw_evaluation_jaad-all}
%     \end{subfigure}
%     %\vspace{4mm}
%     \begin{subfigure}[b]{\linewidth}
%     \centering
%         \caption{JAAD\textsubscript{beh}}
%         \includegraphics[width=0.95\linewidth]{images/jaad_beh}
%         \label{fig:lw_evaluation_jaad-beh}
%     \end{subfigure}
%     \caption{Effect of using the $\mathcal{L}_w$ loss}
%     \label{fig:lw_evaluation}
% \end{figure}

% \begin{figure}[!ht]
%     \centering
%     \includegraphics[width=\linewidth]{images/plot.pdf}
%     \caption{Effect of using the $\mathcal{L}_w$ loss}
%     \label{fig:lw_evaluation}
% \end{figure}

\begin{table*}[!ht]
\centering
  %\resizebox{15cm}{!}{%
  \begin{adjustbox}{width=\linewidth,center}
  \setlength{\tabcolsep}{8pt}
     \begin{tabular}{l c c c c c c c c c c c c c c c}
        \toprule
        & \multicolumn{15}{c}{\textbf{PIE}}\\
         \cmidrule{2-16}
         & \multicolumn{3}{c}{$4$~$s$} & & \multicolumn{3}{c}{$3$~$s$} & &
          \multicolumn{3}{c}{$2$~$s$} & &
          \multicolumn{3}{c}{$1$~$s$}\\
          \cmidrule{2-4} \cmidrule{6-8} \cmidrule{10-12} \cmidrule{14-16}
         & Acc & AUC & F1 & & 
         Acc & AUC & F1 & &
         Acc & AUC & F1 & &
         Acc & AUC & F1\\
         \cmidrule{2-16}
         PCPA\textsuperscript{\ref{foot:reproducibility}}~\cite{Kotseruba2021BenchmarkFE} & $0.76$ & $0.75$ & $0.62$ & & $0.77$ & $0.76$ & $0.63$ & & $0.83$ & $0.84$ & $0.73$ & & $0.86$ & $0.85$ & $0.77$\\
         R-LSTM~\cite{Furnari2019WhatWY} & $0.75$ & $0.64$ & $0.48$ & & $0.75$ & $0.66$ & $0.50$ & & $0.76$ & $0.66$ & $0.51$ & & $0.76$ & $0.67$ & $0.52$\\
         RU-LSTM~\cite{Furnari2019WhatWY} & $0.77$ & $0.76$ & $0.63$ & & $0.80$ & $0.79$ & $0.68$ & & $0.85$ & $0.82$ & $0.74$ & & \textbf{0.88} & $0.85$ & $0.79$\\
         \cdashline{1-16}\noalign{\vskip 0.5ex}
         TAMformer (ours) & \textbf{0.78} & \textbf{0.77} & \textbf{0.65} & & \textbf{0.81} & \textbf{0.81} & \textbf{0.7} & & \textbf{0.87} & \textbf{0.84} & \textbf{0.76} & & \textbf{0.88} & \textbf{0.88} & \textbf{0.8}\\
         \cmidrule{2-16}
         & \multicolumn{15}{c}{\textbf{JAAD}$_{all}$}\\
         \cmidrule{2-16}
         PCPA\textsuperscript{\ref{foot:reproducibility}}~\cite{Kotseruba2021BenchmarkFE} & $0.75$ & $0.75$ & $0.50$ & & $0.74$ & $0.76$ & $0.53$ & & $0.72$ & $0.75$ & $0.52$ & & $0.76$ & $0.79$ & $0.55$\\
         R-LSTM~\cite{Furnari2019WhatWY} & $0.83$ & $0.74$ & $0.54$ & & $0.86$ & $0.73$ & $0.58$ & & $0.85$ & $0.73$ & $0.57$ & & $0.87$ & $0.77$ & $0.62$\\
         RU-LSTM~\cite{Furnari2019WhatWY} & $0.84$ & \textbf{0.76} & \textbf{0.57} & & \textbf{0.87} & $0.78$ & \textbf{0.64} & & $0.85$ & $0.76$ & $0.59$ & & $0.86$ & $0.78$ & $0.62$\\
         \cdashline{1-16}\noalign{\vskip 0.5ex}
         TAMformer (ours) & \textbf{0.85} & $0.75$ & $0.56$ & & $0.86$ & \textbf{0.79} & \textbf{0.64} & & \textbf{0.89} & \textbf{0.82} & \textbf{0.68} & & \textbf{0.89} & \textbf{0.82} & \textbf{0.7}\\
         \cmidrule{2-16}
         & \multicolumn{15}{c}{\textbf{JAAD}$_{beh}$}\\
         \cmidrule{2-16}
         PCPA\textsuperscript{\ref{foot:reproducibility}}~\cite{Kotseruba2021BenchmarkFE} & $0.54$ & $0.51$ & $0.62$ & & $0.47$ & $0.46$ & $0.54$ & & $0.49$ & $0.45$ & $0.61$ & & $0.45$ & $0.52$ & $0.63$\\
         R-LSTM~\cite{Furnari2019WhatWY} & $0.67$ & $0.64$ & $0.74$ & & $0.70$ & $0.64$ & $0.78$ & & $0.66$ & $0.62$ & $0.75$ & & $0.65$ & $0.60$ & $0.75$\\
         RU-LSTM~\cite{Furnari2019WhatWY} & \textbf{0.72} & \textbf{0.67} & \textbf{0.79} & & $0.72$ & $0.64$ & \textbf{0.81} & & $0.69$ & $0.62$ & $0.78$ & & $0.70$ &
         $0.63$ & $0.79$\\
         \cdashline{1-16}\noalign{\vskip 0.5ex}
         TAMformer (ours) & $0.68$ & $0.62$ & $0.77$ & & \textbf{0.73} & \textbf{0.68} & $0.80$ & & \textbf{0.73} & \textbf{0.7} & \textbf{0.79} & & \textbf{0.74} & \textbf{0.69} & \textbf{0.81}\\
         \bottomrule
    \end{tabular}
    %}
    \end{adjustbox}
    \caption{Performance at different anticipation times $[4-1]$s}
    \label{tab:overtime}
\end{table*}

%\smallskip
\noindent 
\textbf{Ablation Experiments.} In Table~\ref{tab:time_scale}, we evaluate the effect of processing input at different time scales in the model. Three approaches are tested: single and fixed scales ($30$ FPS and $10$ FPS), multi-scale (SlowFast [$10$ FPS-$30$ FPS]), and our adaptive scale. As noticed, scaling down can improve performance by discarding much redundant information. Almost better performance can be achieved by applying the SlowFast multi-scaling that allows the model to benefit better from all available information. Yet, allowing the model to choose where to look should be the best option concerning the reported results. Table~\ref{tab:model_variants} compares the model's different variants; increasing the training samples and applying the $\mathcal{L}_r$ loss allow for the best performance. Additionally, Fig.~\ref{fig:lw_evaluation} illustrates the effect of applying the $\mathcal{L}_r$ loss on all anticipation times, where a noticeable increase in the F1-score is present, especially at early anticipation times on the JAAD dataset.

\begin{table}[t!]
\centering
  \resizebox{8cm}{!}{%
     \begin{tabular}{l c c c c c c c c}
        \toprule
        &\multicolumn{3}{c}{\textbf{JAAD}$_{all}$}
        && \multicolumn{3}{c}{\textbf{JAAD}$_{beh}$}\\
         \cmidrule{2-4} \cmidrule{6-8}
         &Acc & AUC & F1 & & 
         Acc & AUC & F1\\
         \cmidrule{2-4} \cmidrule{6-8}
         $30$ FPS & $0.84$ & \underline{0.78} & $0.6$ & & $0.63$ & $0.51$ & \underline{0.77}\\
         $10$ FPS & $0.86$ & \textbf{0.79} & $0.63$ & & \underline{0.64} & $0.56$ & $0.76$\\
         SlowFast & \textbf{0.88} & $0.77$ & $0.64$ & & \textbf{0.67} & \textbf{0.62} & $0.75$\\
         Adaptive & \underline{0.87} & \underline{0.78} & \textbf{0.64} & & \textbf{0.67} & \underline{0.58} & \textbf{0.78}\\
         \bottomrule
    \end{tabular}
    }
    \caption{Effect of Time Scale}
    \label{tab:time_scale}
\end{table}

\begin{figure}[!ht]
    \centering
    \includegraphics[width=\linewidth]{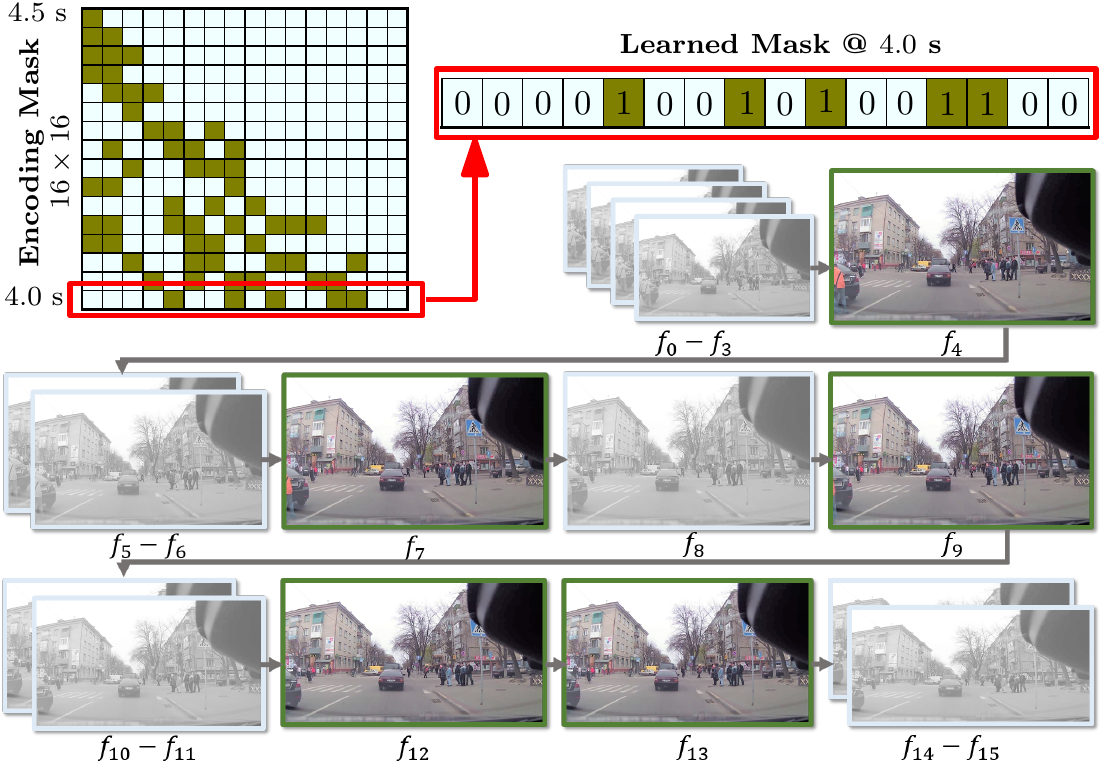}
    \caption{Qualitative example of a learned attention mask}
    \label{fig:qualitative_example}
\end{figure}

\smallskip
\noindent 
\textbf{Qualitative Results.} Fig.~\ref{fig:qualitative_example} is an example of a learned mask and the corresponding input images. For illustration, only the $0.5$s encoding mask is shown. The model chooses a different set of history frames at each time step that should maximize the information and minimize the redundancy at the corresponding time step. For example, at $t_a = 4$s, the model uses only $5$ frames from the available $16$ frames. Given the raw images, we observe much redundancy, yet some differences in the chosen images by the model.

\begin{table}[!hb]
\centering
  \begin{adjustbox}{width=\linewidth,center}
  \setlength{\tabcolsep}{2pt}
     \begin{tabular}{c c c c c c c c c c c c c c c c}
        \toprule
        \multirow{2}{*}{TAS}&\multirow{2}{*}{DI}&\multirow{2}{*}{$\mathcal{L}_r$}
        &&\multicolumn{3}{c}{\textbf{JAAD}$_{all}$}
        && \multicolumn{3}{c}{\textbf{JAAD}$_{beh}$}&&\multicolumn{3}{c}{\textbf{PIE}}\\
         \cmidrule{5-7} \cmidrule{9-11} \cmidrule{13-15}
         &&&&Acc & AUC & F1 & & 
         Acc & AUC & F1 & &
         Acc & AUC & F1\\
         \cmidrule{1-3} \cmidrule{5-7} \cmidrule{9-11} \cmidrule{13-15}
          %\xmark & \xmark & \xmark & $0.84$ & $0.78$ & $0.6$ & & $0.63$ & $0.51$ & $0.77$ & & \textbf{-} & \textbf{-} & \textbf{-}\\
         \cmark & \xmark & \xmark && $0.87$ & $0.78$ & $0.64$ & & $0.67$ & $0.58$ & $0.78$ & & \textbf{0.88} & $0.85$ & $0.78$\\
         \cmark & \cmark & \xmark && \textbf{0.88} & $0.79$ & $0.65$ & & $0.69$ & $0.68$ & $0.75$ && \textbf{-} & \textbf{-} & \textbf{-}\\
         \cmark & \cmark & \cmark && \textbf{0.88} & \textbf{0.83} & \textbf{0.68} & & \textbf{0.73} & \textbf{0.69} & \textbf{0.8} && \textbf{0.88} & \textbf{0.86} & \textbf{0.79}\\
         \bottomrule
    \end{tabular}
    \end{adjustbox}
    \caption{Models Variants (\textbf{DI} stands for Data Increase)}
    \label{tab:model_variants}
\end{table}

%%% conclusion
%\vspace{-5mm}
\section{Conclusions and Future Work}
\label{sec:conclusion}
In this work, we propose a multi-modality transformer-based model that can learn attention masks adaptively to measure the temporal sequence's correspondences. We applied a new loss function to minimize the gap in performance between early anticipation times and the closest one to the anticipated action. The experiments demonstrate the proposed model's out-performance, which can reach $+2\%$ F1 on PIE and $+5\%$ F1 on JAAD, in the $[2-1]$s range. Similarly, TAMformer surpasses at early anticipation times, mainly on PIE. Yet, our model suffers a drop in performance at early anticipation times on JAAD. Thus, our future work will focus on achieving robust performance at all anticipation times.

%%% references
\ninept
\vfill\pagebreak
\bibliographystyle{IEEEbib}
\bibliography{ref}

\end{document}